\documentclass[11pt]{article}

\usepackage[preprint]{acl}

\usepackage{times}
\usepackage{latexsym}

\usepackage[T1]{fontenc}

\usepackage[utf8]{inputenc}

\usepackage{microtype}

\usepackage{inconsolata}

\usepackage{graphicx}
\usepackage{amsmath}
\usepackage{amssymb}
\usepackage{amsfonts}
\usepackage{multirow}
\usepackage{adjustbox}
\usepackage{booktabs}
\usepackage[table]{xcolor}
\usepackage{siunitx}
\usepackage{xspace}
\newcommand{\MethodName}{RADAR\xspace}
\title{RADAR: Reasoning as Discrimination with Aligned Representations for LLM-based Knowledge Graph Reasoning}

\author{
Bo Xue\thanks{Equal contribution. $\dagger$ Corresponding author.},
Yuan Jin\footnotemark[1],
Luoyi Fu,
Jiaxin Ding$^\dagger$,
Xinbing Wang \\
Shanghai Jiao Tong University, Shanghai, China \\
\texttt{\{sappho\_x, jiaxinding\}@sjtu.edu.cn}
}

\begin{document}
\maketitle
\begin{abstract}
Knowledge graph reasoning (KGR) infers missing facts, with recent advances increasingly harnessing the semantic priors and reasoning abilities of Large Language Models (LLMs). However, prevailing generative paradigms are prone to memorize surface-level co-occurrences rather than learning genuine relational semantics, limiting out-of-distribution generalization. To address this, we propose \MethodName, which reformulates KGR from generative pattern matching to discriminative relational reasoning. We recast KGR as discriminative entity selection, where reinforcement learning enforces relative entity separability beyond token-likelihood imitation. Leveraging this separability, inference operates directly in representation space, ensuring consistency with the discriminative optimization and bypassing the generation-induced hallucinations. Across four benchmarks, \MethodName achieves 5–6\% relative gains on link prediction and triple classification over strong LLM baselines, while increasing task-relevant mutual information in intermediate representations by 62.9\%, indicating more robust and transferable relational reasoning. 
\end{abstract}

\section{Introduction}

Knowledge graphs (KGs) represent facts as structured triples and support applications such as semantic search and question answering~\cite{chen2020knowledge,huang2019knowledge,wang2024knowledge}. However, real-world KGs are inevitably incomplete, limiting their effectiveness in downstream use~\cite{shen2022comprehensive}. Knowledge graph reasoning (KGR) addresses this by inferring missing facts from observed triples. Classical embedding-based methods exploit graph structure effectively, but often struggle under sparsity and have limited capacity to express fine-grained relational semantics beyond local topology~\cite{pujara2017sparsity,yao2019kg}. Consequently, recent work has shifted toward large language models (LLMs), harnessing their rich semantic priors and reasoning capabilities to mitigate these limitations~\cite{zhang2024making,yao2025exploring,li2024cosign,li2024contextualization}.

\begin{figure}[t]
    \centering
    \includegraphics[width=0.95\linewidth]{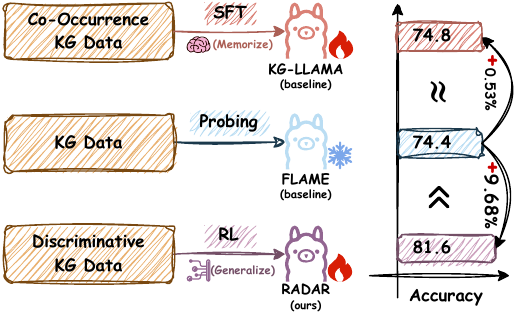}
    \caption{FB15K-237N triple classification with a shared LLaMA backbone: co-occurrence-based SFT yields only marginal gains over the frozen baseline, whereas \MethodName substantially improves accuracy by optimizing discriminative relational reasoning.}
    \label{fig:intro}
    \vspace{-4mm}
\end{figure}

\begin{figure*}[t]
    \centering
    \includegraphics[width=0.95\linewidth]{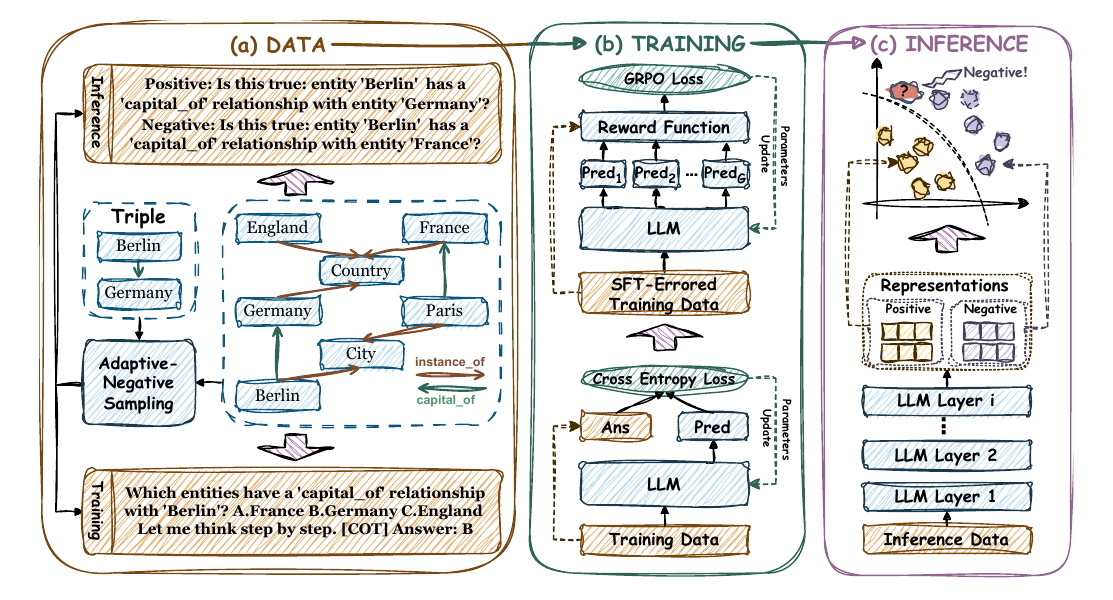}
    \caption{\MethodName, a data-to-inference alignment framework for discriminative KGR.}
    \label{fig:arc}
    \vspace{-4mm}
\end{figure*}

A central requirement of KGR is \emph{generalization}: the ability to answer queries involving entity--relation combinations unseen during training~\cite{liu2021kg,bordes2013translating}. Yet prevailing LLM-based paradigms formulate KGR as sequence modeling, optimizing next-token likelihood over serialized textual triples. This formulation encourages a shortcut: models minimize loss by exploiting surface co-occurrences between entity names and relations rather than learning relation-conditioned validity~\cite{zhang2024co,kang2023impact,ju2024investigating}. Such behavior yields strong in-distribution performance but fails under distribution shift where those co-occurrence statistics no longer hold. Crucially, this reliance on co-occurrence shortcuts is further exacerbated by supervised fine-tuning (SFT), as cross-entropy objectives reinforce token-level imitation over relational reasoning~\cite{chu2025sft,lv2025towards}. As illustrated in Figure~\ref{fig:intro}, co-occurrence-based SFT yields only marginal gains over the frozen baseline, highlighting the necessity for a paradigm shift from generative pattern matching to relational reasoning.

To address co-occurrence-driven shortcuts, we propose \MethodName(Figure~\ref{fig:arc}), which fundamentally shifts KGR from generative pattern matching to discriminative relational reasoning. By recasting KGR as discriminative entity selection, we redirect the learning signal from token-level imitation to the relative separability of ground truth against hard distractors, rendering co-occurrence shortcuts insufficient~\cite{si2022language}. This reformulation pivots KGR from local token-level prediction to global entity discrimination within a discrete candidate space, where Reinforcement Learning (RL) naturally aligns with optimizing global discrete outcomes and effectively resolving the objective mismatch inherent in token-level SFT~\cite{chu2025sft}. Operationalizing this shift necessitates staged training, where RL exploits SFT-grounded structure to amplify reward-driven contrast between correct entities and distractors, transforming the objective from likelihood-based imitation to generalizable discrimination~\cite{lv2025towards}. As a consequence of optimizing for relative entity separability, relational information is encoded in discriminative latent representations rather than through autoregressive decoding~\cite{zou2023representation}. Accordingly, we perform inference directly in representation space, aligning inference with the discriminative signals induced during training and avoiding generation-induced hallucinations.

Across four KGR benchmarks, \MethodName consistently achieves an average 5–6\% relative improvement over strong LLM-based baselines on link prediction and triple classification. Ablations show that the gains come from a tightly integrated design that aligns what the model is trained to optimize, what it is allowed to rely on, and what is ultimately extracted for reasoning. Furthermore, our proposed information-theoretic probe reveals that \MethodName increases task-relevant mutual information in intermediate representations by 62.9\% on average, consistent with improved inductive robustness and domain transfer. Our contributions are threefold:

\begin{itemize}
    \item We expose co-occurrence shortcuts as a fundamental bottleneck in LLM-based KGR and recast KGR as a discriminative relational reasoning problem, moving beyond autoregressive pattern matching to achieve robust out-of-distribution generalization.
    \item We introduce \MethodName, a data-to-inference alignment framework that restructures supervision, optimization, and inference around discrete entity separation, inducing relational semantics to emerge as discriminative structures within the representation space and bypassing surface-level co-occurrence shortcuts.
    \item Extensive evaluations across four benchmarks demonstrate superior performance over strong baselines, while information-theoretic quantification validates that \MethodName substantially enhances task-relevant mutual information to achieve robust inductive generalization.
\end{itemize}

\section{Methods}

\subsection{Setup}

A knowledge graph $\mathcal{G}=\{\mathcal{E},\mathcal{R},\mathcal{T}\}$ is a collection of an entity set $\mathcal{E}$, a relation set $\mathcal{R}$, and a triple set $\mathcal{T}$. Each triple is denoted as $(h,r,t)\in \mathcal{T}$, where $h,t\in\mathcal{E}$ represent the entities and $r\in\mathcal{R}$ denotes the relation. Let $\ell(\cdot)$ denote the function that maps each entity or relation to its textual sequence for LLM processing.

In this work, we focus on two standard KGR tasks: triple classification and link prediction. Triple classification determines whether a triple $(h, r, t)$ is valid. Link prediction aims to identify the missing entity in an incomplete triple $(h, r, ?)$.

\subsection{Task Reformulation for Discriminative Selection}\label{sec:data}

Prevailing generative paradigms~\cite{saxena2022sequence,yao2025exploring} optimizing serialized triples (e.g., "$\ell(h)$ $\ell(r)$ $\ell(t)$") inherently biases models toward minimizing loss via surface-level entity-relation co-occurrences rather than relational validity~\cite{kang2023impact,ju2024investigating}. To redirect this bias, we recast KGR as discriminative entity selection within a constrained candidate space. This reformulation fundamentally alters the learning signal: the model is compelled to explicitly discriminate ground truth from distractors, thereby forcing the optimization trajectory toward inducing discriminative latent structures governed by relational validity, rendering co-occurrence shortcuts insufficient for objective minimization.

\begin{figure}[htb]
    \centering
    \includegraphics[width=1\linewidth]{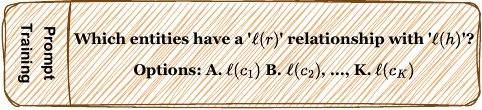}
    \caption{Training prompt.}
    \label{fig:prompt_train}
    \vspace{-1mm}
\end{figure}

Formally, given a query triple $(h, r, ?)$, we define $\mathcal{E}_{\text{gt}}(h, r) = \{t \mid (h, r, t) \in \mathcal{T}\}$ as the ground-truth tail entity set. We construct a  candidate set $\mathcal{C}(h,r) = \{c_1, c_2, \ldots, c_K\}$, where $\mathcal{C}(h,r) = \mathcal{E}_{pos}(h,r) \cup \mathcal{E}_{neg}(h,r)$, $\mathcal{E}_{pos}(h,r) \subseteq \mathcal{E}_{\text{gt}}(h, r)$ denotes positive entities and $\mathcal{E}_{neg}(h,r)\subset \mathcal{E} \setminus \mathcal{E}_{\text{gt}}(h, r)$ denotes negative entities. Each training instance is formatted as shown in Figure~\ref{fig:prompt_train}.

The model is trained to output the option labels corresponding to entities in $\mathcal{E}_{pos}(h,r)$.

\paragraph{Hierarchical Task Difficulty.} To progressively challenge the model's relational reasoning capabilities, we design a hierarchical framework along two dimensions: answer cardinality and negative sample hardness.

For answer cardinality, we construct two task variants: (1)~\emph{single-answer}: $|\mathcal{E}_{pos}(h,r)| = 1$; (2)~\emph{variable-answer}: $|\mathcal{E}_{pos}(h,r)| \ge 1$, where the number of correct entities is not given a priori, encompassing both one-to-one and one-to-many relations and requiring cardinality determination.

For negative sample hardness, we stratify negative candidates based on their plausibility given the query. We use a pre-trained knowledge graph embedding (KGE) model to score all potential entities and partition $\mathcal{E}_{neg}(h,r)$ into three tiers based on plausibility. (1)~\emph{Tier 1 (Easy)}: lowest-scoring entities that are structurally implausible as tails for the query; (2)~\emph{Tier 2 (Medium)}: randomly sampled entities; (3)~\emph{Tier 3 (Hard)}: highest-scoring entities that are highly confusable with ground-truth answers, requiring fine-grained relational reasoning.

\subsection{Optimization for Discriminative Relational Reasoning}\label{sec:training}

To overcome the pattern matching, we develop a two-stage training paradigm tightly coupled with discrete entity selection formulation. While SFT grounds output structure, its cross-entropy objective inherently favors token-level imitation over relational reasoning~\cite{lv2025towards}. This mismatch necessitates RL, which pivots the learning signal from token likelihood to global outcome validity~\cite{chu2025sft}. Crucially, the  reformulation in Section~\ref{sec:data} yields explicit, verifiable rewards, enabling stable RL and promoting genuine relational discrimination over statistical mimicry.

\paragraph{Stage I: Supervised Fine-Tuning.} We construct chain-of-thought (CoT) reasoning traces~\cite{wei2022chain} for each training instance, where each trace verifies semantic compatibility of candidate entities with the query $(h,r,?)$ before producing the final answer set (see Appendix~\ref{sec:appendix_training} for details). Given a training dataset $\mathcal{D}_{\text{train}}$, where each instance consists of a discriminative prompt $x$ and a target sequence comprising the CoT reasoning and the final answer. The model is fine-tuned using the standard next-token prediction objective.

\paragraph{Stage II: Reinforcement Learning.} We apply Group Relative Policy Optimization (GRPO)~\cite{shao2024deepseekmath}, leveraging group-normalized advantages for stable policy updates, to enhance generalization on challenging entity-relation compositions.

Let $\hat{y}$ denote the model output for input $x$, and let $\hat{\mathcal{A}}(\hat{y}) \subseteq \mathcal{C}(h,r)$ denote the set of entities extracted from $\hat{y}$. To amplify reasoning signals on instances where SFT fails, we construct an error-focused dataset $\mathcal{D}_{\text{error}}$ by evaluating the SFT model on $\mathcal{D}_{\text{train}}$ and retaining only instances where $\hat{\mathcal{A}}(\hat{y}) \neq \mathcal{E}_{pos}(h,r)$. 

To guide learning on $\mathcal{D}_{\text{error}}$ while accommodating variable-cardinality predictions, we define a composite reward $R(x, \hat{y})$:

\begin{equation}		
\label{eq:reward}
R(x, \hat{y}) = \alpha \cdot R_{\text{fmt}}(\hat{y}) + (1-\alpha) \cdot R_{\text{acc}}(x, \hat{y})
\end{equation}

\noindent where $\alpha$ is the weighting coefficient, $R_{\text{fmt}}(\hat{y}) \in \{0, 1\}$ verifies format adherence, and $R_{\text{acc}}(x, \hat{y})$ measures answer accuracy. For each query $(h, r, ?)$ encoded in $x$, the accuracy reward is:

\begin{equation}		
\label{eq:reward_acc}
R_{\text{acc}}(x, \hat{y}) = \frac{2 \cdot |\hat{\mathcal{A}}(\hat{y}) \cap \mathcal{E}_{pos}(h,r)|}
{|\hat{\mathcal{A}}(\hat{y})| + |\mathcal{E}_{pos}(h,r)|}
\end{equation}

\noindent This F1-based formulation provides credit assignment for partially correct predictions, which is essential for variable-answer instances. We optimize the policy to maximize the expected reward using GRPO, with details provided in Appendix~\ref{sec:appendix_training}.

\subsection{Representation-based Inference} \label{sec:extraction}

By explicitly discriminating against distractors, the task reformulation and staged training shape the model's internal representations toward relational separability. Autoregressive decoding evaluates relational plausibility indirectly through next-token likelihood, introducing a mismatch between token-level generation and the entity-level discriminative separability induced during training~\cite{zou2023representation,orgad2024llms}. We therefore perform inference directly in representation space, enabling faithful extraction of discriminative relational knowledge and ensuring that the inference mechanism is intrinsically aligned with the discriminative learning signals optimized during training.

For triple classification, we extract the model's internal representations to assess triple plausibility. We construct prompt $\text{PT}_{\text{cls}}(h,r,t)$ as follows:

\begin{figure}[htb]
    \centering
    \includegraphics[width=1\linewidth]{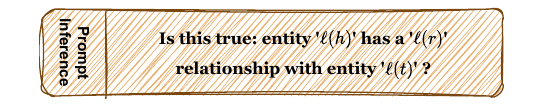}
    \caption{Inference prompt.}
    \label{fig:prompt_inference}
    \vspace{-3mm}
\end{figure}

We extract the hidden state at the final token position from an intermediate layer $l$, as intermediate representations have been shown to better preserve factual knowledge than final-layer outputs~\cite{skean2025layer}:

\begin{equation}
\mathbf{z}^{(l)}(h, r, t) = (\text{LLM}^{(l)}\left(\text{PT}_{\text{cls}}(h, r, t)\right))_{T},    
\end{equation}

\noindent where $\mathbf{z}^{(l)}(h, r, t)\in \mathbb{R}^{d}$ denotes the extracted hidden state, $\text{LLM}^{(l)}(\cdot)$ represents output of the first $l$ layers of the fine-tuned model, $(\cdot)_{T}$ extracts the representation at position $T$ (the final token), and $d$ is the hidden dimension.

We then train a binary classifier $f_\phi: \mathbb{R}^d \to [0, 1]$ that maps the extracted representation to a plausibility score $s(h, r, t) = f_\phi(\mathbf{z}^{(l)}(h, r, t))$. Training data consists of positive samples from ground-truth triples and negative samples constructed by corrupting tail entities~\cite{bordes2013translating}. We optimize $f_\phi$ using binary cross-entropy loss and implement $f_\phi$ as a two-layer MLP, which maps $\mathbf{z}^{(l)}$ through a hidden layer of dimension $d_v$ to produce the binary classification score.

For link prediction, where the goal is to identify the missing tail entity in $(h, r, ?)$, we adopt a retrieval-then-reranking approach~\cite{li2025retrieval}. Given the large entity set size, exhaustively scoring all entities is computationally prohibitive. We therefore leverage a lightweight pre-trained KGE model to retrieve the top-$n$ structurally plausible candidates ($n \ll |\mathcal{E}|$), then rerank them using our learned classifier $f_\phi$. For each retrieved candidate $e_i$, we compute $s(h, r, e_i) = f_\phi(\mathbf{z}^{(l)}(h, r, e_i))$ and rank candidates by this score.

\subsection{Task-Adaptive Information Quantification}\label{sec:smi}

To quantify discriminative relational signals internalized during our reformulated KGR training, we propose a task-adaptive mutual information measure defined over internal representations $\mathbf{z}^{(l)}$. While standard sliced mutual information (SMI) relies on random projections that treat all directions equally, this indiscriminate averaging can dilute task‑relevant information in LLM representations~\cite{goldfeld2021sliced,wongso2023using}. In contrast, we derive projections from the learned probing classifier $f_\phi$ to target the most discriminative subspace, concentrating information estimation on directions that encode relational separability induced by discriminative KGR training.

Since $f_\phi$ is optimized to distinguish valid triples, its weight matrix encodes discriminative directions in the representation space. For $N$ samples with representations $\mathbf{Z}^{(l)} \in \mathbb{R}^{d \times N}$, we define task-adaptive projections using the first layer of $f_\phi$:
\begin{equation}
    \mathbf{V} = \text{PReLU}(\mathbf{W}_1 \mathbf{Z}^{(l)} + \mathbf{b}_1) \in \mathbb{R}^{d_v \times N},
\end{equation}
where $\mathbf{W}_1 \in \mathbb{R}^{d_v \times d}$ and $\mathbf{b}_1 \in \mathbb{R}^{d_v}$ are the trained parameters. Each row $\mathbf{v}_i \in \mathbb{R}^{N}$ of $\mathbf{V}$ corresponds to the representations projected onto a dimension optimized for KGR.

Let $Y \in \{0,1\}^N$ denote the ground-truth validity labels. We then quantify the task-specific information encoded in the representation space by averaging the mutual information between the projected feature activations and the labels:
\begin{equation}
    \mathcal{I}_{\text{task}} = \frac{1}{d_v} \sum_{i=1}^{d_v} \hat{I}(\mathbf{v}_{i}; Y),
\end{equation}
where $\hat{I}$ is mutual information computed via the KSG method~\cite{kraskov2004estimating}. This metric quantifies the KGR-relevant information within the representations that is captured by the discriminative features of $f_\phi$, enabling comparison across training strategies (details in Appendix \ref{sec:appendix_smi}).

\section{Experiment}\label{sec:experiment}

Our experiments address the following research questions:
\textbf{RQ1:} Does \MethodName achieve strong and consistent performance across standard KGR benchmarks and tasks?
\textbf{RQ2:} How do the core design components of \MethodName individually and synergistically contribute to performance and generalization?
\textbf{RQ3:} Can \MethodName achieve robust inductive generalization to unseen entities and transfer relational knowledge to domain-related tasks?

\subsection{Experiment Settings}

\textbf{Datasets.} 
We evaluate \MethodName on four benchmarks across two tasks. For link prediction, we use FB15K-237~\cite{schlichtkrull2018modeling} and FB15K-237N~\cite{lv2022pre}. For triple classification, we employ WN18RR~\cite{dettmers2018convolutional}, UMLS~\cite{yao2019kg}, and FB15K-237N. Dataset details are provided in Appendix~\ref{sec:datasets}.

\textbf{Baselines.} 
We compare \MethodName against two categories of baselines:
1) Knowledge Graph Embedding models, which map entities and relations into low-dimensional vector spaces, including TransE~\cite{bordes2013translating}, DistMult~\cite{yang2014embedding}, ComplEx~\cite{trouillon2016complex}, RotatE~\cite{sun2019rotate}, ConvE~\cite{dettmers2018convolutional}, and TuckER~\cite{balavzevic2019tucker}.
2) Language Model-based methods, which leverage the textual semantics of KGs. These include KG-BERT~\cite{yao2019kg}, MTL-KGC~\cite{kim2020multi}, KG-S2S~\cite{chen2022knowledge}, SimKGC~\cite{wang2022simkgc}, iGT~\cite{luo2025gltw}, CSPromp-KG~\cite{chen2023dipping}, COSIGN~\cite{li2024cosign}, KG-LLAMA~\cite{yao2025exploring}, CD~\cite{li2024contextualization}, LLAMA-ICL, Structure IT, KoPA~\cite{zhang2024making} and FLAME~\cite{xue2024unlock}. Among these, KG-LLAMA and FLAME are the most directly comparable baselines because they rely solely on LLaMA's intrinsic capabilities without external structural embeddings, in contrast to hybrid approaches such as KoPA. Further details on these baselines are provided in Section~\ref{sec:related}.

\textbf{Evaluation Metrics.} 
For link prediction, we report the Mean Reciprocal Rank (MRR) and Hits@$k$ ($k=1,3,10$) following standard protocols. 

\textbf{Implementation Details.}
Unless otherwise noted, we adopt the variable-answer setting with Tier 2 negative sampling (random negatives) for all experiments. To ensure fair comparison with KG-LLAMA and FLAME, we adopt LLaMA~\cite{touvron2023llama} as the backbone model across all configurations. For link prediction inference and negative sample stratification in data construction (Section~\ref{sec:data}), we employ TuckER~\cite{balavzevic2019tucker} as the KGE model to score candidate entities. Additional implementation details are provided in Appendix~\ref{sec:appendix_add}.

\begin{table*}[t] 
    \centering
    \caption{Link prediction performance (MRR and Hits@k) on FB15K-237 and FB15K-237N. Best scores are in \textbf{bold} and second-best are \underline{underlined}.}

    \label{tab:link_prediction}
    \small
    \renewcommand{\arraystretch}{1.0} 
    \setlength{\tabcolsep}{7.7pt}
    
    \begin{tabular}{llcccccccc}
        \toprule
        \multirow{2}{*}{\textbf{Paradigm}} & \multirow{2}{*}{\textbf{Model}} & \multicolumn{4}{c}{\textbf{FB15K-237}} & \multicolumn{4}{c}{\textbf{FB15K-237N}} \\
        \cmidrule(lr){3-6} \cmidrule(lr){7-10}
        & & \textbf{MRR} & \textbf{H@1} & \textbf{H@3} & \textbf{H@10} & \textbf{MRR} & \textbf{H@1} & \textbf{H@3} & \textbf{H@10} \\
        \midrule
        
        \multirow{7}{*}{\shortstack[l]{Embedding\\-based}}
        & TransE   & 0.279 & 0.198 & 0.376 & 0.441 & 0.255 & 0.152 & 0.301 & 0.459 \\
        & DistMult & 0.281 & 0.199 & 0.301 & 0.446 & 0.209 & 0.143 & 0.234 & 0.330 \\
        & ComplEx  & 0.278 & 0.194 & 0.297 & 0.450 & 0.249 & 0.180 & 0.276 & 0.380 \\
        & RotatE   & 0.338 & 0.241 & 0.375 & 0.533 & 0.279 & 0.177 & 0.320 & 0.481 \\
        & ConvE    & 0.312 & 0.225 & 0.341 & 0.497 & 0.273 & 0.192 & 0.305 & 0.429 \\
        & TuckER   & 0.347 & 0.253 & 0.382 & 0.536 & 0.301 & 0.217 & 0.332 & 0.463 \\
        \midrule
        
        \multirow{9}{*}{\shortstack[l]{Language\\Model\\-based}}
        & KG-BERT    & 0.237 & 0.169 & 0.260 & 0.427 & 0.203 & 0.139 & 0.201 & 0.403 \\
        & MTL-KGC    & 0.267 & 0.172 & 0.298 & 0.458 & 0.241 & 0.160 & 0.284 & 0.430 \\
        & SimKGC     & 0.333 & 0.246 & 0.362 & 0.510 & 0.372 & 0.289 & 0.402 & \underline{0.534} \\
        & KG-S2S     & 0.336 & 0.257 & 0.373 & 0.498 & 0.353 & 0.282 & 0.385 & 0.495 \\
        & CSPromp-KG & 0.358 & 0.269 & 0.393 & 0.538 & 0.360 & 0.281 & 0.395 & 0.511 \\
        & CD         & --    & --    & --    & --    & 0.372 & 0.288 & 0.410 & 0.530 \\
        & COSIGN     & \underline{0.368} & \textbf{0.315} & \textbf{0.434} & 0.520 & \underline{0.394} & \textbf{0.355} & \underline{0.457} & 0.526 \\
        & KG-LLAMA   & 0.238 & 0.165 & 0.272 & 0.423 & --    & --    & --    & --    \\

        \midrule
        
        \rowcolor{gray!10} 
        \multicolumn{2}{c}{\textbf{\MethodName}} & \textbf{0.377} & \underline{0.273} & \underline{0.421} & \textbf{0.579} & \textbf{0.415} & \underline{0.301} & \textbf{0.476} & \textbf{0.633} \\
        \bottomrule
    \end{tabular}
\end{table*}
\begin{table}[t]
    \centering
    \caption{Triple classification accuracy on WN18RR, FB15K-237N, and UMLS. Best scores are in \textbf{bold} and second-best are \underline{underlined}.}
    \label{tab:triple_classification}
    \small
    \renewcommand{\arraystretch}{1.0} 
    \setlength{\tabcolsep}{2.0pt}
    
    \begin{tabular}{llccc}
        \toprule
        \textbf{Paradigm} & \textbf{Model} & \textbf{WN18RR} & \textbf{FB15K-237N} & \textbf{UMLS} \\
        \midrule
        
        \multirow{4}{*}{\shortstack[l]{Embedding\\-based}}
        & TransE   & 88.4 & 69.7 & 84.5 \\
        & DistMult & 85.1 & 58.7 & 86.4 \\
        & ComplEx  & 84.1 & 65.7 & 87.1 \\
        & RotatE   & 88.2 & 68.5 & 87.7 \\
        \midrule
        
        \multirow{6}{*}{\shortstack[l]{Language\\Model\\-based}}
        & KG-BERT            & 91.6 & 56.0 & 77.3 \\
        & LLAMA-ICL          & 50.2 & 59.2 & 55.5 \\
        & KG-LLAMA           & 92.1 & 74.8 & 85.8 \\
        & Structure IT & 92.7 & 76.4 & 89.9 \\
        & FLAME              & 93.8 & 74.4 & 86.6 \\
        & KoPA               & \underline{94.9} & \underline{77.7} & \textbf{92.6} \\
        \midrule

        \rowcolor{gray!10} 
        \multicolumn{2}{c}{\textbf{\MethodName}} & \textbf{95.3} & \textbf{81.6} & \underline{91.7} \\
        \bottomrule
    \end{tabular}
    \vspace{-3mm}
\end{table}

\subsection{Main Results (RQ1)}

Table~\ref{tab:link_prediction} reports link prediction results of \MethodName on FB15K-237 and FB15K-237N. Specifically, \MethodName attains the best MRR across both datasets, outperforming all KGE-based and language model-based baselines, while ranking first or second on all Hits@k metrics. On FB15K-237, \MethodName reaches an MRR of 0.377, corresponding to a 2.4\% relative improvement over the previous best model COSIGN. Notably, \MethodName demonstrates substantial improvements on the more challenging FB15K-237N dataset, achieving a 5.3\% improvement in MRR over COSIGN. 

To complement link prediction, we further evaluate triple classification on WN18RR, FB15K-237N, and UMLS as shown in Table~\ref{tab:triple_classification}.  \MethodName achieves the best accuracy on WN18RR and FB15K-237N, and is second on UMLS (91.7\%), slightly below KoPA (92.6\%). KoPA leverages external structural embeddings, whereas \MethodName operates in the LLM-only setting. Focusing on the LLM-only setting with the same LLaMA backbone, \MethodName substantially improves over KG-LLAMA and FLAME, yielding an average relative gain of 6.1\% and 5.7\% across the three datasets, respectively, suggesting reduced reliance on entity–relation co-occurrence shortcuts and stronger generalization in LLaMA-based KGR.

\begin{table}[htb]
    \centering
    \caption{Triple classification accuracy on FB15K-237N under controlled ablations of task difficulty: answer cardinality and negative hardness.}
    \label{tab:task_complexity}
    \setlength{\tabcolsep}{2pt} 
    \begin{small}
    \begin{tabular}{lc c lc}
        \toprule
        \multicolumn{2}{c}{\textbf{Answer Cardinality}} & & \multicolumn{2}{c}{\textbf{Negative Hardness}} \\
        \multicolumn{2}{c}{\textit{\scriptsize (Fixing Negatives: Tier 2)}} & & \multicolumn{2}{c}{\textit{\scriptsize (Fixing Cardinality: Variable)}} \\
        \cmidrule(r){1-2} \cmidrule(l){4-5} 
        Setting & Accuracy & & Setting & Accuracy \\
        \midrule
        Single-answer & 79.9 & & Tier 1 \textit{(Easy)} & 80.2 \\
        Variable-answer & 81.6 & & Tier 2 \textit{(Medium)} & 81.6 \\
         &  & & Tier 3 \textit{(Hard)} & 81.0 \\
        \bottomrule
    \end{tabular}
    \end{small}
    \vspace{-3mm}
\end{table}

Finally, Table~\ref{tab:task_complexity} studies task variants on FB15K-237N by varying answer cardinality and negative hardness. The variable-answer setting performs better than single-answer, suggesting that exposing multiple correct tails for one-to-many relations provides richer supervision for relation-conditioned discrimination. Under the variable-answer setting, Tier-2 is marginally better than Tier-3 and Tier 1, and overall performance differences across tiers remain small, motivating our default choice of Tier-2 in subsequent experiments. See Appendix~\ref{sec:appendix_ablation} for further analysis on negative hardness.

Taken together, these results show that \MethodName delivers consistent gains across link prediction and triple classification under a unified discriminative training-and-inference pipeline, outperforming prior LLM-only baselines while remaining competitive against methods that explicitly fuse KG structural embeddings.

\subsection{Ablation Results (RQ2)}\label{sec:ablation}

We ablate \MethodName along three axes—data formulation (serialization vs. discrimination), training objective (sft vs. two-stage training), and inference (generation vs. representation extraction)—using four variants named by \textsc{Data–Train–Infer}. \emph{Serial-SFT-Generation} follows the KG-LLaMA-style generative setup, fine-tuning on serialized triples (e.g., Is this true: $\ell(h)\ \ell(r)\ \ell(t)$?). \emph{Serial-SFT-Extraction} keeps the same serialized SFT setup but replaces decoding with representation extraction-based inference (Section~\ref{sec:extraction}), isolating the effect of inference. \emph{Discriminative-SFT-Extraction} replaces serialization with the discriminative candidate-selection formulation (Section~\ref{sec:data}), which reduces entity–relation co-occurrence shortcuts while keeping SFT-only training and extraction inference, isolating the impact of task formulation. \emph{Discriminative-Full-Extraction} represents our full framework with discriminative formulation, two-stage full training (Section~\ref{sec:training}), and extraction inference, isolating the effect of two-stage training. 

\begin{table}[htb]
    \centering
    \caption{Ablation across task formulation, training objective, and inference strategy on triple classification accuracy (WN18RR, FB15k-237N, UMLS).}
    \label{tab:module_ablation}
    \small
    \setlength{\tabcolsep}{2pt} 
    \begin{tabular}{lccc} 
    \toprule
    \textbf{Method} & \textbf{WN18RR} & \textbf{15K-237N} & \textbf{UMLS} \\
    \midrule
    Serial-SFT-Generation            & 0.921 & 0.748 & 0.858 \\
    Serial-SFT-Extraction            & 0.925 & 0.753 & 0.861 \\
    Discriminative-SFT-Extraction    & 0.945 & 0.793 & 0.898 \\
    Discriminative-Full-Extraction & \textbf{0.953} & \textbf{0.816} & \textbf{0.917} \\
    \bottomrule
    \end{tabular}
\end{table}

Table~\ref{tab:module_ablation} quantifies the contribution of each design axis in \MethodName. First, replacing autoregressive decoding with representation-based inference (\emph{Serial-SFT-Extraction}) yields consistent gains over generation (\emph{Serial-SFT-Generation}), suggesting that internal representations capture relational information more reliably than autoregressive outputs. Second, reformulating KGR as a discriminative task leads to substantial performance improvements. Specifically, \emph{Discriminative-SFT-Extraction} improves over \emph{Serial-SFT-Extraction} by an average of 3.9\% across benchmarks. These improvements align with our hypothesis that the discriminative formulation mitigates entity–relation co-occurrence shortcuts and promotes reasoning over relational semantics. Incorporating RL refinement yields further gains, with \emph{Discriminative-Full-Extraction} achieving an average relative improvement of 6.5\% over the standard serial SFT generation baseline. Together, these results show that effective relational reasoning in LLM-based KGR requires a tightly integrated design across task formulation, optimization, and inference. In this system, outcome-based optimization does not merely refine supervised learning, but works in concert with the discriminative setup and representation-based inference to resolve non-trivial challenges in aligning what the model is trained to optimize, what it is allowed to rely on, and what knowledge is ultimately extracted for reasoning.

\begin{figure}[htb]
    \centering
    \includegraphics[width=1.0\linewidth]{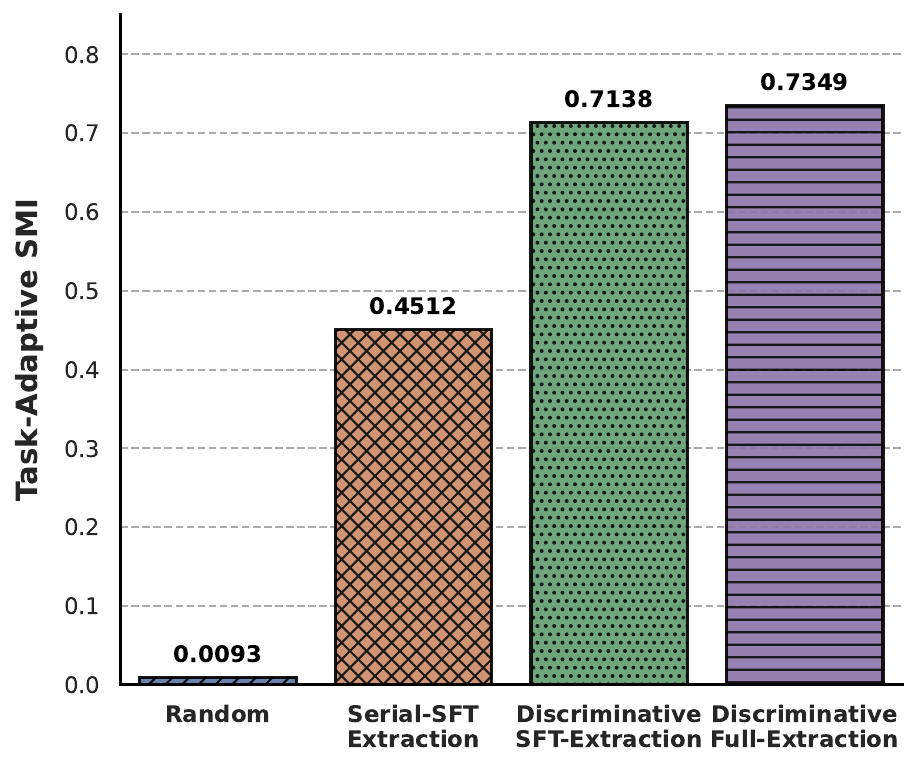}
    \caption{Task-Adaptive SMI of intermediate representations across different methods on WN18RR.}
    \label{fig:smi}
\end{figure}

To better interpret the performance gains observed in the ablation study, we measure how each design choice affects the amount of KGR-relevant information encoded in intermediate representations using task-adaptive mutual information $\mathcal{I}_{\text{task}}$. As shown in Figure~\ref{fig:smi}, \MethodName yields a 62.9\% average relative increase in $\mathcal{I}_{\text{task}}$ over the serialized SFT baseline on WN18RR (please refer to Appendix~\ref{sec:appendix_smi} for corresponding analyses on other datasets). Discriminative reformulation establishes a representation space suitable for relational discrimination, while reinforcement learning further aligns this space with outcome-level relational correctness beyond supervised likelihood. Together, these results indicate that the gains of \MethodName stem from systematic changes in how task-relevant information is organized in the representation space.

\begin{table}[htb]
  \centering
  \small
  \renewcommand{\arraystretch}{1.4} 
  \setlength{\tabcolsep}{2pt}
  \caption{Link prediction performance on FB15K-237N across different LLM backbones.}
  \label{tab:model_ablation}
  \begin{tabular}{lcccc}
    \toprule
    \textbf{Method} & \textbf{MRR} & \textbf{H@1} & \textbf{H@3} & \textbf{H@10} \\
    \midrule
    \multicolumn{5}{l}{\textit{\textbf{LLaMA-7B}}~\cite{touvron2023llama}} \\ 
    \quad Serial-SFT-Extraction          & 0.365 & 0.229 & 0.438 & 0.584 \\
    \quad Discriminative-Full-Extraction & \textbf{0.415} & \textbf{0.301} & \textbf{0.476} & \textbf{0.633} \\
    \addlinespace[0.5em] 
    \multicolumn{5}{l}{\textit{\textbf{Pythia-6.9B}}~\cite{biderman2023pythia}} \\
    \quad Serial-SFT-Extraction          & 0.353 & 0.259 & 0.401 & 0.616 \\
    \quad Discriminative-Full-Extraction & \textbf{0.384} & \textbf{0.267} & \textbf{0.437} & \textbf{0.635} \\
    \addlinespace[0.5em]
    \multicolumn{5}{l}{\textit{\textbf{Qwen3-8B}}~\cite{yang2025qwen3}} \\
    \quad Serial-SFT-Extraction          & 0.402 & 0.287 & 0.442 & 0.622 \\
    \quad Discriminative-Full-Extraction & \textbf{0.426} & \textbf{0.312} & \textbf{0.499} & \textbf{0.641} \\
    \bottomrule
  \end{tabular}
\end{table}

We evaluate the architectural robustness of \MethodName by examining whether its relative improvements are preserved across diverse LLM backbones.  As shown in Table~\ref{tab:model_ablation}, \MethodName consistently outperforms the corresponding Serial-SFT-Extraction baselines across all evaluated architectures. Specifically, on Qwen3-8B, \MethodName yields an relative improvement of 6.0\% in MRR and 8.7\% in Hits@1 over the Serial-SFT-Extraction baseline. These results indicate that \MethodName is robust to backbone choice, yielding consistent performance improvements across diverse LLM architectures.

\subsection{Additional Results (RQ3)}

\begin{figure}[htb]
    \vspace{-3mm}
    \centering
    \includegraphics[width=0.99\linewidth]{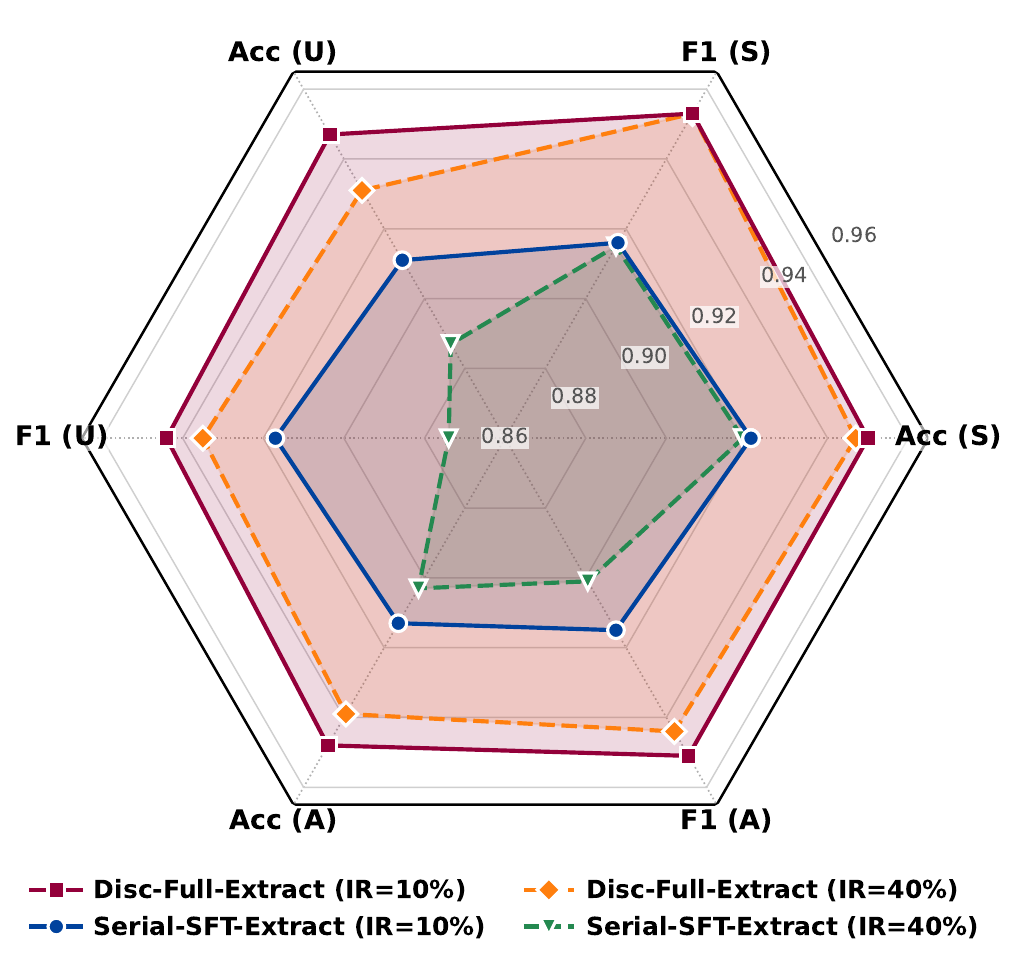}
    \caption{Inductive generalization performance for triple classification on WN18RR under varying inductive rates (IR). We report accuracy on seen (S), unseen (U), and overall (A) test triples.}
    \label{fig:radar}
    \vspace{-3mm}
\end{figure}

We evaluate inductive generalization under an entity-disjoint setting in which a subset of entities is completely excluded from training, directly testing the model’s ability to reason beyond observed entity identities~\cite{teru2020inductive}. Following~\cite{chen2022meta}, we define an inductive rate (IR) as the proportion of entities excluded from training. Given IR=$\rho$, a fraction $\rho$ of entities is designated as inductive, and all training triples involving them are removed. We stratify test triples into seen (S) where all entities appeared in the training set, unseen (U) involving at least one inductive entity, and (A)  denoting the aggregate set. This setting presents a stringent generalization challenge, as inductive entities are never observed during training.

\emph{Discriminative-Full-Extraction} consistently outperforms the Serial-SFT baseline across all inductive rates, with particularly pronounced gains on unseen triples. At IR=40\%, the baseline exhibits a sharp degradation on unseen triples, while our method remains substantially more stable. This gap supports our hypothesis that serialization-based SFT encourages reliance on entity co-occurrence statistics, which fails when such correlations are unavailable. In contrast, \MethodName encourages the model to internalize relation-conditioned distinction patterns that are decoupled from specific entity identities. As a result, the learned relational criteria can be consistently applied to unseen entities at inference time, indicating that the components interact in a principled and complementary manner rather than functioning as isolated heuristics.

To evaluate whether knowledge learned through KGR generalizes beyond the training task, we conduct zero-shot evaluations on MMLU ~\cite{hendrycks2020measuring} subjects aligned with the UMLS domain, following the protocol of \cite{zhang2024making}. As shown in Table~\ref{tab:mmlu}, KGR training on UMLS leads to an average performance gain of 11.6\% across five subjects. These gains suggest that our KGR training encourages relation-conditioned and compositional reasoning that generalizes to domain-adjacent tasks, rather than reliance on memorized triples.

\begin{table}[htb]
  \centering
  \caption{Zero-shot performance on domain-aligned MMLU subjects after UMLS-based KGR training.}
  \label{tab:mmlu}
    \small
    \renewcommand{\arraystretch}{1.4} 
    \setlength{\tabcolsep}{5.0pt}
  \begin{tabular}{lcc}
    \toprule
    \textbf{Subjects} & \textbf{w/o Training} & \textbf{w/ Training} \\
    \midrule
    Clinical              & 0.340  & \textbf{0.455} \\
    College Medicine      & 0.341 & \textbf{0.397} \\
    High School Biology   & 0.313 & \textbf{0.439} \\
    High School Chemistry & 0.271 & \textbf{0.355} \\
    Medical Genetics      & 0.290  & \textbf{0.490}  \\
    \bottomrule
  \end{tabular}
\end{table}

\section{Conclusion}

In this work, we introduce \MethodName, a framework that fundamentally realigns LLM-based KGR from autoregressive pattern matching to discriminative relational reasoning. By unifying task reformulation, optimization, and inference toward discriminative entity selection, our approach suppresses memorization of surface statistics in favor of robust relational semantics.  Validated by information-theoretic analysis and extensive benchmarks, \MethodName not only achieves superior generalization but also establishes a principled route to more reliable and generalizable KGR with LLMs.

\clearpage

\section{Limitations}

To balance large-scale inference efficiency, \MethodName adopts a retrieve-then-rerank paradigm, introducing an inherent recall bottleneck. Additionally, our discriminative alignment via reinforcement learning incurs higher training costs than standard fine-tuning. Future work will focus on optimizing training efficiency and exploring retriever-free LLM reasoning mechanisms.

\bibliography{custom}

\clearpage

\appendix

\section{Related Work} \label{sec:related}

Knowledge graph reasoning aims to infer missing triples from incomplete KGs~\cite{tang2024rule}. Classical methods learn embedding-based scoring functions over graph topology,  including translational models (e.g., TransE~\cite{bordes2013translating}, RotatE~\cite{sun2019rotate}), semantic matching models (e.g., DistMult~\cite{yang2014embedding}, ComplEx~\cite{trouillon2016complex}, TuckER~\cite{balavzevic2019tucker}), and convolutional models (e.g., ConvE~\cite{dettmers2018convolutional}). While effective at capturing local  structural patterns, these methods struggle with data sparsity.

Language model-based methods address these limitations by incorporating textual semantics.  Early encoder-only approaches, including KG-BERT \cite{yao2019kg} and LASS \cite{shen2022joint}, reformulate KGC as sequence classification. Later work enhances this paradigm through multi-task learning \cite{kim2020multi} and contrastive objectives \cite{wang2022simkgc}. Generative methods such as KG-S2S \cite{chen2022knowledge} and KGT5 \cite{saxena2022sequence} adopt generative paradigms with encoder-decoder architectures. CSProm-KG \cite{chen2023dipping} introduces conditional soft prompts to balance structural and textual knowledge. Recent advances leverage decoder-only LLMs for KGC. KoPA \cite{zhang2024making} combines structural embeddings with Alpaca for triple classification, KG-LLAMA \cite{yao2025exploring} frames KGR as instruction-following question answering with parameter-efficient fine-tuning and FLAME leverages frozen LLM representations with task-specific KGC classifiers.

However, existing LM-based methods primarily rely on supervised fine-tuning over data with explicit entity-relation co-occurrences, which biases models toward memorizing surface-level statistical patterns rather than relational reasoning. In contrast, \MethodName co-designs the task formulation, training objective, and inference mechanism to prioritize generalization over surface-form memorization in KGR.

\section{Datasets and Metrics Details}\label{sec:datasets}

\begin{table}[htb]
    \centering
    \caption{Dataset statistics.}
    \label{tab:dataset}
    \small
    \setlength{\tabcolsep}{3pt}
    \begin{tabular}{lccccc} 
    \toprule
    \textbf{Dataset} & $\vert \mathbf{\mathcal{E}} \vert$ & \textbf{$\vert \mathbf{\mathcal{R}} \vert$} & \textbf{\# Train} & \textbf{\# Valid} & \textbf{\# Test} \\
    \midrule
    FB15K-237   & 14,541 & 237 & 272,115 & 17,535 & 20,466 \\
    FB15K-237N  & 13,104 & 93  & 87,282  & 7,041  & 8,226  \\
    WN18RR      & 40,943 & 11  & 86,835  & 6,068  & 6,268  \\
    UMLS        & 135   & 46  & 5,216   & 1,304  & 1,322  \\
    \bottomrule
    \end{tabular}
\end{table}

We evaluate the performance of \MethodName on four benchmark datasets, covering two distinct tasks: link prediction and triple classification. The statistics of these datasets are summarized in Table~\ref{tab:dataset}. For link prediction, we utilize FB15K-237, a subset of Freebase with inverse relations removed to prevent data leakage, and FB15K-237N, a modified version that removes mediator nodes and introduces hard negatives to increase difficulty. For triple classification, we employ WN18RR, a subset of WordNet derived from WN18 by removing inverse relations; UMLS, a medical semantic network consisting of biomedical concepts and their semantic relations; and the aforementioned FB15K-237N.

For link prediction, we report the Mean Reciprocal Rank (MRR) and Hits@$k$ ($k=1,3,10$) following standard protocols. MRR is the average of the reciprocal ranks of the correct entities, while Hits@$k$ measures the proportion of correct entities ranked in the top-$k$. For triple classification, we report accuracy. To ensure a fair evaluation, the test sets for triple classification are constructed with an equal number of positive and negative triples.

\section{Additional Implementation Details}\label{sec:appendix_add}

For each query $(h,r,?)$, we construct a discriminative candidate set $\mathcal{C}(h,r)$ with a fixed size of $K=4$, containing both positive and negative tail entities as described in Section~\ref{sec:data}. 

Entity descriptions are sourced from \citet{xie2016representation} for FB15K-237 and FB15K-237N, synset definitions from \citet{yao2019kg} for WN18RR, and the original dataset for UMLS. Both Stage I SFT and Stage II GRPO are implemented using LoRA for parameter-efficient fine-tuning, with all backbone parameters kept frozen. For representation-based inference, we consistently extract hidden states from an intermediate layer with index $l=15$, following prior work showing that intermediate representations better preserve relational knowledge. 

For link prediction, exhaustively scoring the entire entity set $\mathcal{E}$ is computationally prohibitive for LLMs. We therefore adopt a retrieve-then-rerank strategy~\cite{li2025retrieval,wei2024kicgpt}, utilizing a lightweight pre-trained TuckER model to retrieve the top-$n=15$ candidates, which are subsequently reranked by our learned classifier. For MRR calculation, should the ground truth entity fall outside this retrieved top-$n$ set, we use the rank provided by the retriever. While this imposes a recall bottleneck, it balances large-scale inference efficiency with precision, offering a more rigorous evaluation than ranking against randomly sampled negatives~\cite{yao2025exploring}. Crucially, this reliance on retrieval is exclusive to link prediction task. Our triple classification results (Table~\ref{tab:triple_classification}) and the core analytical experiments in Section~\ref{sec:experiment} are conducted without retrieval, directly validating the intrinsic discriminative reasoning capabilities of \MethodName.

\begin{table}[h]
    \centering
    \caption{Link classification performance on FB15K-237N under varying parameter $n$.}
    \label{tab:ablation_n}
    \vspace{0.2cm} 
    \small
    \begin{tabular}{cc} 
        \toprule
        \textbf{Parameter} $n$ & \textbf{MRR} \\
        \midrule
        15 & \textbf{0.415} \\ 
        20 & 0.402 \\
        25 & 0.393 \\
        \bottomrule
    \end{tabular}
\end{table}

\section{Additional Details on Task-Adaptive Information Quantification}\label{sec:appendix_smi}

Section~\ref{sec:smi} introduces a task-adaptive mutual information metric for quantifying how much KGR-relevant signal is encoded in intermediate representations. Here we provide additional motivation for this design and explain why standard SMI may be suboptimal in our setting.

Standard SMI approximates $I(X;Y)$ by averaging mutual information over random one-dimensional projections. However, in LLM representations, task-relevant information often concentrates in low-dimensional subspaces~\cite{hu2022lora}, and random projections dilute this signal by uniformly sampling directions regardless of their informativeness. This leads to noisier estimates with weaker task alignment. Moreover, restricting to linear projections can be suboptimal when the relevant structure is expressed through nonlinear feature interactions.

To address these limitations, We leverage the trained probe $f_\phi$ (Section~\ref{sec:extraction}) to define task-aligned projections. Given $N$ samples with representations $\mathbf{Z}^{(l)}\in\mathbb{R}^{d\times N}$, we compute the probe's first-layer activations:

\begin{equation}
\mathbf{V} = \text{PReLU}(\mathbf{W}_1 \mathbf{Z}^{(l)} + \mathbf{b}_1)\in \mathbb{R}^{d_v \times N},    
\end{equation}

where $\mathbf{W}_1 \in \mathbb{R}^{d_v \times d}$ is the learned weight matrix, $\mathbf{b}_1\in \mathbb{R}^{d_v}$ is the bias vector (with broadcasting across samples), and $\mathbf{V} \in \mathbb{R}^{d_v \times N}$ is the matrix of transformed hidden representations. Crucially, since $\mathbf{W}_1$ is optimized via supervised learning to minimize classification loss on KGR, each row of $\mathbf{W}_1$ defines a projection direction that is implicitly optimized for label discriminability.

This learned transformation differs from random projections in two respects. First, the PReLU nonlinearity enables modeling of nonlinear feature interactions inherent to contextualized embeddings. Second, supervised optimization shapes $\mathbf{W}_1$ to emphasize label-informative subspaces while suppressing task-irrelevant dimensions. As a result, the hidden layer of $f_\phi$ defines a compact, task-aligned projection space that preserves KGR-relevant information, leading to more focused and better aligned mutual information estimates than random projections.

\begin{figure}[htb]
    \centering
    \includegraphics[width=1.0\linewidth]{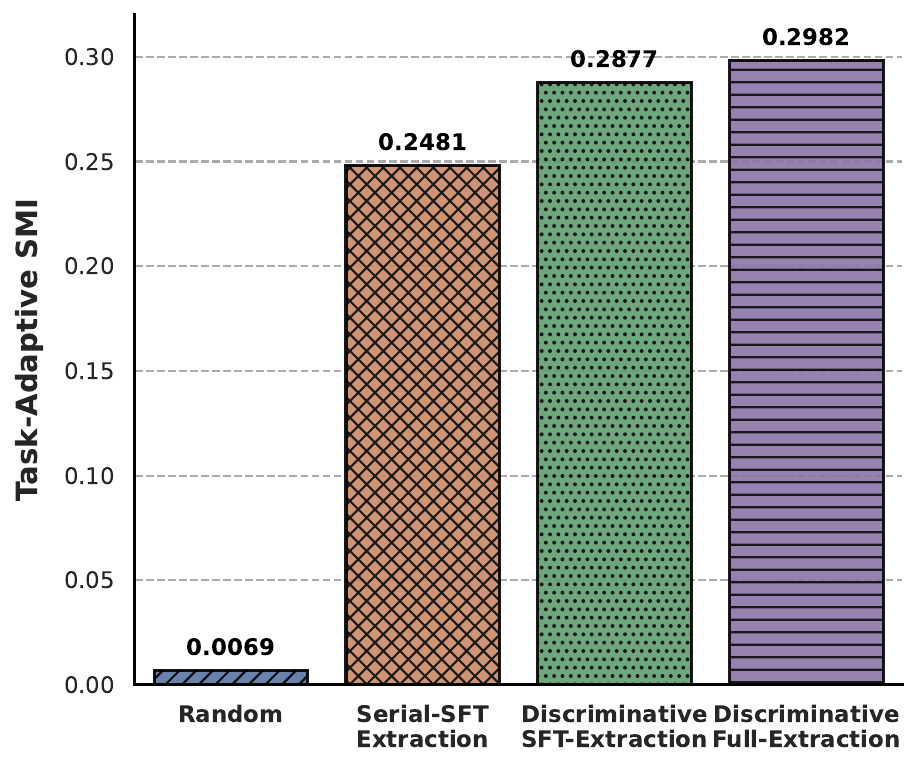}
    \caption{Task-Adaptive SMI of intermediate representations across different methods on FB15K-237N.}
    \label{fig:smi}
\end{figure}

\section{Additional Details on Training}\label{sec:appendix_training}

\paragraph{Stage I: Supervised Fine-Tuning.}
In Stage I, we perform supervised fine-tuning using structured CoT rationales to guide relational reasoning. For each entity, we use a long-form textual description $d(\cdot)$. Prompts are constructed using a fixed template that presents the semantics of the head entity and all candidate entities, followed by the answer selection instruction. The prompts are shown in Figure~\ref{fig:prompt_cot}:

\begin{figure}[htb]
    \centering
    \includegraphics[width=1\linewidth]{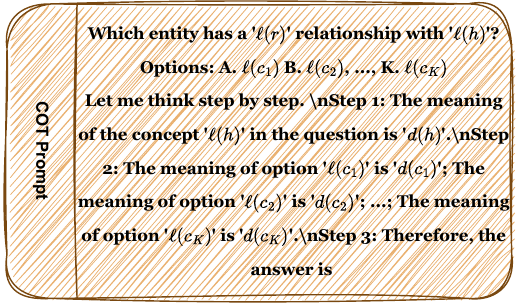}
    \caption{COT prompt.}
    \label{fig:prompt_cot}
    \vspace{-1mm}
\end{figure}

Formally, given a training dataset $\mathcal{D}_{\text{train}}$, where each instance consists of a discriminative prompt $x$ constructed from $(h,r,\mathcal{C}(h,r))$ and a target sequence $y$ comprising the CoT reasoning and the final answer. We optimize the model parameters $\theta$ using the standard next-token prediction objective:

\begin{equation}		
\label{eq:sft}
\mathcal{L}_{\text{SFT}} = -\mathbb{E}_{(x,y) \sim \mathcal{D}_{\text{train}}} \left[ \sum_{t=1}^{|y|} \log p_\theta(y_t \mid x, y_{<t}) \right]
\end{equation}

\paragraph{Stage II: Reinforcement Learning.} 
We optimize the policy using GRPO, which updates the model by comparing the relative quality of sampled responses for each input. For each query, the model samples a group of $G$ candidate responses $\{\hat{y}_i\}_{i=1}^G$, each scored by $R(x, \hat{y}_i)$. Rewards are normalized within each group to compute advantages that guide policy updates. The objective is:

\vspace{-5mm}
\begin{equation}
\label{eq:grpo_1}
\begin{aligned}
\mathcal{J}_{\text{GRPO}}(\theta) = 
&\mathbb{E}_{x\sim\mathcal{D}_{\text{error}}, \{\hat{y}_i\} \sim \pi_{\theta_{\text{old}}}} 
\Bigg[
\frac{1}{G} \sum_{i=1}^G L_i^{clip} \\
&\phantom{\mathbb{E}_{x\sim\mathcal{D}_{\text{error}},}} \
- \beta \, \mathbb{D}_{\text{KL}}(\pi_{\theta} \,||\, \pi_{\text{ref}})
\Bigg],
\end{aligned}
\end{equation}
\vspace{-6mm}

\begin{equation}		
\label{eq:grpo_2}
L_i^{clip} = 
\min \left( w_i A_i, 
\text{clip}(w_i, 1 - \epsilon, 1 + \epsilon) A_i \right),
\end{equation}

\noindent where $w_i = \frac{\pi_\theta(\hat{y}_i \mid x)}{\pi_{\theta_{\text{old}}}(\hat{y}_i \mid x)}$, $\pi_{\theta_{\text{old}}}$ is the policy before the update, $\pi_{\text{ref}}$ is the reference policy (the initial SFT model), $\epsilon$ and $\beta$ are hyperparameters controlling update clipping and KL regularization, and $A_i$ denotes the normalized advantage of each candidate within its group based on the reward $R(x, \hat{y}_i)$.

\section{Additional Ablation Results}\label{sec:appendix_ablation}

\begin{figure}[htb]
    \centering
    \includegraphics[width=1.0\linewidth]{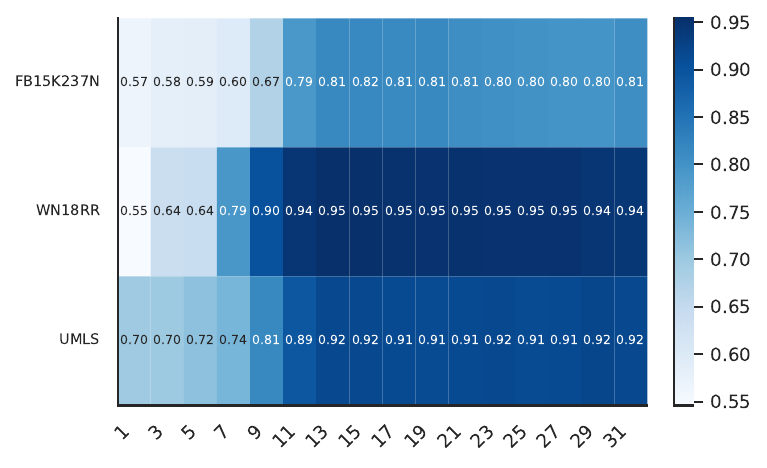}
    \caption{Layer-wise triple classification accuracy of \MethodName across datasets. The horizontal axis denotes the layer depth in LLaMA, and the vertical axis corresponds to different datasets.}
    \label{fig:layer}
\end{figure}

Figure~\ref{fig:layer} shows that as the layer depth increases, the effectiveness of hidden states for triple classification exhibits a unimodal trend, with performance first improving and then degrading. Hidden states from intermediate and upper layers generally yield higher prediction accuracy than those from lower layers. This observation is consistent with prior findings~\cite{geva2020transformer}, which show that Transformer-based language models encode knowledge hierarchically across layers, with higher layers progressively integrating information from lower-level representations. During pretraining, the lower layers may not have stored certain knowledge, thus failing to produce hidden states informative for triple classification. Furthermore, layers closest to the output head tend to underperform relative to middle layers in triple classification accuracy. This phenomenon may be attributed to the capacity of intermediate layers to strike an optimal balance between semantic abstraction and information retention. While lower layers contain raw input noise and upper layers converge toward generic token-prediction objectives, intermediate layers distill task-salient features that are most conducive to discriminative relational reasoning~\cite{jiang2024large,zou2023representation}.

\begin{table}[t]
    \centering
    \caption{Triple classification accuracy on UMLS under varying parameter $K$ of Discriminative-SFT-Extraction. The best result is highlighted in bold.}
    \label{tab:ablation_k}
    \vspace{0.2cm} 
    \small
    \begin{tabular}{cc} 
        \toprule
        \textbf{Parameter} $K$ & \textbf{Accuracy} \\
        \midrule
        3 & 0.892 \\
        4 & \textbf{0.898} \\ 
        5 & 0.887 \\
        6 & 0.893 \\
        \bottomrule
    \end{tabular}
\end{table}

Finally, we investigate the impact of the candidate set size $K$ in Table~\ref{tab:ablation_k}. While increasing $K$ raises task difficulty—potentially necessitating more robust representations to distinguish the ground truth from a larger pool of distractors—we observe that performance peaks at $K=4$ and subsequently plateaus. This phenomenon parallels our findings on negative hardness in Table~\ref{tab:task_complexity}, where escalating difficulty from Tier 2 (Medium) to Tier 3 (Hard) yields no significant performance gain.

These results collectively suggest that the primary driver of our method's effectiveness is not the severity of the classification task, but the mechanism of discrimination itself. The critical factor is the paradigm shift: by constraining the LLM to select among discrete options, we successfully sever the reliance on generative co-occurrence shortcuts and force the activation of relational reasoning circuits. Once this discriminative mode is engaged, the model learns to isolate relational semantics; further increasing the complexity of the candidate space (via larger $K$ or harder negatives) offers diminishing returns and may introduce optimization noise rather than stronger learning signals.
\end{document}